\documentclass[10pt,twocolumn,letterpaper]{article}

 \usepackage{cvpr}              %

\newcommand{\mysection}[1]{\vspace{2pt}\noindent\textbf{#1}}

\global\long\def\upperBoundT{T}%

\definecolor{cvprblue}{rgb}{0.21,0.49,0.74}
\usepackage[pagebackref,breaklinks,colorlinks,allcolors=cvprblue]{hyperref}
\usepackage{multirow}

\usepackage{natbib}

\def\paperID{13} %
\def\confName{CVsports}
\def\confYear{2026}

\title{Towards Athlete Fatigue Assessment from Association Football Videos}

\author{
{Xavier Bou}$^{1,2}$\quad
{Nathan Correger}$^{3,4}$\quad 
{Alexandre Cloots}$^{3}$\quad
{Cédric Gavage}$^{3,5}$\quad 
{Silvio Giancola}$^{6}$\quad \\
{Cédric Schwartz}$^{3}$\quad 
{François Delvaux}$^{3}$\quad 
{Rudi Cloots}$^{3}$\quad 
{Marc Van Droogenbroeck}$^{3}$\quad 
{Anthony Cioppa}$^{3}$\\[0.5em]
$^1$Université Paris-Saclay, CNRS, ENS Paris-Saclay \quad
$^2$AMIAD, Pôle Recherche \quad \\
$^3$University of Liège\quad
$^4$CESI\quad 
$^5$HEPL \quad
$^6$KAUST \quad
}

\begin{document}
\maketitle
\begin{abstract}
Fatigue monitoring is central in association football due to its links with injury risk and tactical performance. 
However, objective fatigue-related indicators are commonly derived from subjective self-reported metrics, biomarkers derived from laboratory tests, or, more recently, intrusive sensors such as heart monitors or GPS tracking data. 
This paper studies whether monocular broadcast videos can provide spatio-temporal signals of sufficient quality to support fatigue-oriented analysis. Building on state-of-the-art Game State Reconstruction methods, we extract player trajectories in pitch coordinates and propose a novel kinematics processing algorithm to obtain temporally consistent speed and acceleration estimates from reconstructed tracks. 
We then construct acceleration–speed (A--S) profiles from these signals and analyze their behavior as fatigue-related performance indicators. 
We evaluate the full pipeline on the public SoccerNet-GSR benchmark, considering both 30-second clips and a complete 45-minute half to examine short-term reliability and longer-term temporal consistency. 
Our results indicate that monocular GSR can recover kinematic patterns that are compatible with A--S profiling while also revealing sensitivity to trajectory noise, calibration errors, and temporal discontinuities inherent to broadcast footage. 
These findings support monocular broadcast video as a low-cost basis for fatigue analysis and delineate the methodological challenges for future research.%
\end{abstract}

\begin{figure}
    \centering
    \includegraphics[width=\linewidth]{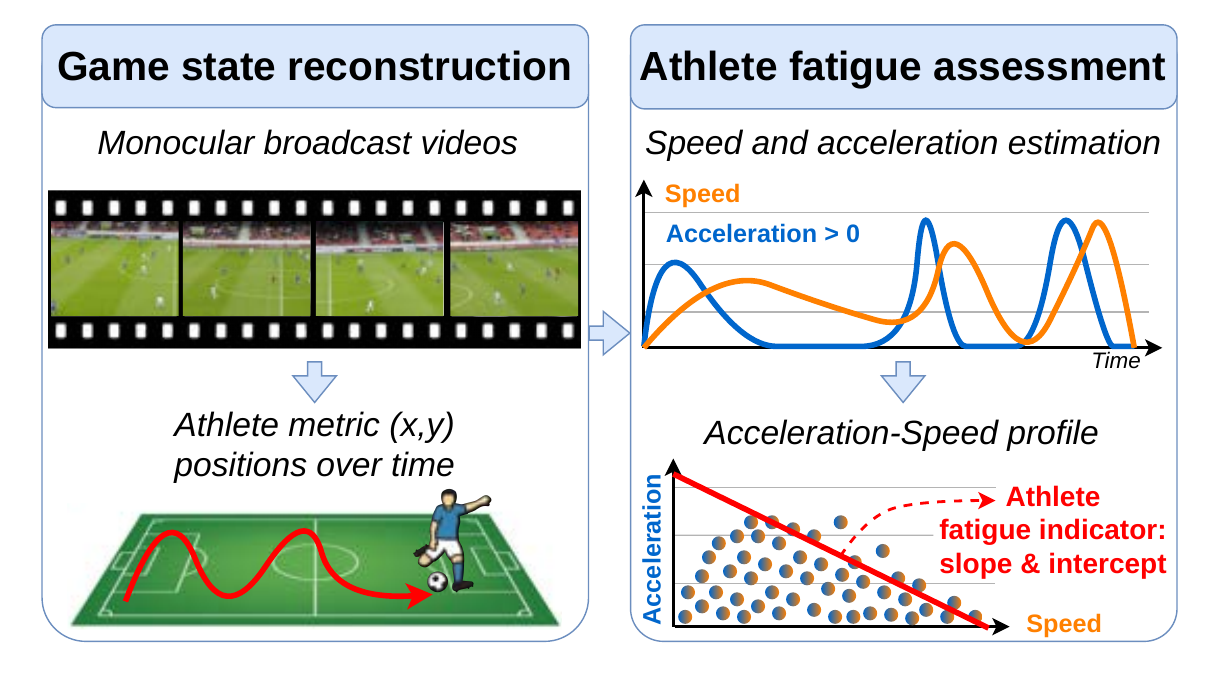}
    \vspace{-2em}
    \caption{\textbf{Overview of the proposed method for athlete fatigue assessment from monocular broadcast football videos.} Left: Game state reconstruction enables the recovery of metric player trajectories by projecting detections into pitch coordinates, resulting in time-evolving $(x,y)$ positional signals for each athlete. Right (top): From these trajectories, temporally consistent speed and acceleration signals are estimated in the metric coordinate system. Right (bottom): Acceleration--Speed (A--S) profiles are constructed by modeling the relationship between instantaneous acceleration and running speed. The slope and intercept parameters of the A--S diagram are analyzed as fatigue-related kinematic indicators, allowing the study of their evolution over match time.}
    \label{fig:graphical_abstract}
    \vspace{-2em}
\end{figure}

\section{Introduction}
\label{sec:intro}
In sports science, fatigue refers to the decline in muscular force or performance of an athlete over time due to acute or accumulated physical and/or mental load~\cite{BestwickStevenson2022Assessment}. Fatigue has been associated with increased injury risk~\cite{Santamaria2010TheEffect} and has been shown to play a fundamental role in tactical dynamics. For example, \citet{daSilva2025TheInterplay} observed that even a small fatigue increase in professional football can lead to a disproportionate loss of spatial control. Consequently, fatigue monitoring is essential to adapt training workloads, prevent injuries, and optimize individual and team performance.

Fatigue can be assessed through a range of physiological and behavioral indicators~\cite{Djaoui2017Monitoring}. Subjective indicators estimate the fatigue levels of athletes by self-reported metrics such as mood or rating of perceived stress and muscular load~\cite{Saw2016Monitoring}. Recent works combined such subjective indicators with GPS information~\cite{Midoglu2024ALargescale} to forecast time to injury in elite female football~\cite{Catterall2026TimetoInjury-arxiv}. In professional settings, these reports are often complemented by objective measurements, including heart rate, blood, or urine sampling~\cite{Djaoui2017Monitoring}. Although biochemical markers provide reliable indicators of physiological strain, they are more costly to obtain and require slower, logistically demanding acquisition procedures. 
Moreover, such measurements are typically obtained before or after match actions, limiting their ability to characterize the temporal evolution of fatigue during a match. To address these constraints, recent studies have explored objective fatigue proxies derived from spatio-temporal performance signals, such as total distance covered or sprint frequency over time~\cite{Pons2021Integrating, Morin2021Individual}. 
In particular, practitioners increasingly rely on acceleration–speed (A--S) profiles, which model the linear relationship between a player's instantaneous acceleration and running speed across match actions~\cite{AlonsoCallejo2023Validity}. These approaches generally rely on proprietary GPS tracking data or expensive multi-camera systems. As a result, reproducibility and scalability remain limited, especially since many existing studies depend on non-public datasets and specialized, proprietary hardware.

In parallel, computer vision methods for sports video understanding have progressed substantially in recent years. Community-driven initiatives such as the SoccerNet challenges~\cite{Giancola2022SoccerNet,Cioppa2024SoccerNet2023Challenge,Cioppa2024SoccerNet2024-arxiv,Giancola2025Deep} have driven research on core perception tasks relevant to football analytics, including player and ball detection~\cite{Cioppa2022Scaling}, tracking~\cite{Cioppa2022SoccerNetTracking,Maglo2022Efficient}, camera calibration~\cite{Magera2024AUniversal, Magera2025BroadTrack}, and jersey number recognition~\cite{Balaji2023Jersey,Li2018Jersey}. More recently, the Game State Reconstruction (GSR) task~\cite{Somers2024SoccerNetGameState} unified these components into a single benchmark designed to extract structured spatio-temporal metric data from broadcast football videos. As single-camera methods trained on public data achieve increasing levels of performance on this benchmark, a natural question arises: can such vision-based pipelines provide sufficiently reliable spatio-temporal indicators to support objective fatigue estimation in professional association football, without relying on GPS data or expensive multi-view acquisition systems?

In this work, we investigate this question by leveraging state-of-the-art monocular Game State Reconstruction (GSR) methods~\cite{Oo2026Broadcast2Pitch,Yang2026SoccerMaster} developed within the SoccerNet framework~\cite{Somers2024SoccerNetGameState,Giancola2025SoccerNet-arxiv}, which enable the automatic extraction of player trajectories from broadcast videos. Building upon these trajectories, we introduce a novel method to compute physically consistent speed and acceleration signals in pitch coordinates, from which acceleration–speed (A--S) profiles are constructed to derive established fatigue-related indicators~\cite{Morin2021Individual}, as illustrated in \cref{fig:graphical_abstract}. 
We evaluate the proposed pipeline on the SoccerNet GSR dataset, including short 30-second clips, as well as on a full 45-minute half game. The analysis focuses on assessing the stability and interpretability of the derived A--S profiles under realistic broadcast conditions.
All components of the framework rely exclusively on publicly available methods and datasets, ensuring complete reproducibility. Our objective is not to provide a definitive physiological quantification of fatigue, but rather to investigate whether monocular pipelines can generate spatio-temporal signals of sufficient quality to support fatigue-related analysis. The study therefore serves as a first step toward single-camera fatigue estimation in association football, identifying both the potential and limitations that must be addressed to enable reliable downstream use.

\mysection{Contributions.} We summarize our contributions as follows. (i) We present the first study investigating athlete fatigue-related indicators automatically derived exclusively from monocular broadcast football videos.
(ii) We introduce a novel method for extracting temporally consistent speed and acceleration signals from player trajectories reconstructed by state-of-the-art game state reconstruction pipelines, enabling the construction of acceleration–speed (A--S) profiles from broadcast data.
(iii) We provide an empirical analysis on the SoccerNet-GSR dataset, including both short clips and a full 45-minute half, examining the feasibility of fatigue-oriented indicator extraction and identifying methodological limitations that affect reliability to drive further research on this topic.

\section{Related works}
\label{sec:related_works}

\subsection{Fatigue in sports science}

In sports science, fatigue is generally described as a multi-factorial phenomenon comprising both peripheral (muscular) and central components. Peripheral fatigue manifests at the level of the muscle fibers, whereas central fatigue is primarily associated with processes in the central nervous system, notably the cortex~\cite{TorneroAguilera2022Central}. Within peripheral fatigue, two mechanisms are commonly distinguished. \emph{Metabolic peripheral fatigue} arises when the energetic demands of intense or prolonged exercise exceed available resources, leading to glycogen depletion and metabolite accumulation (\eg, elevated lactate) within the muscle~\cite{TorneroAguilera2022Central}. In contrast, \emph{non-metabolic peripheral fatigue} is linked to micro-damage within muscle fibers~\cite{Kano2012Mechanisms}, where excessive stretching of the sarcolemma can induce small tears and trigger repair processes that substantially extend recovery time. Recovery timelines typically span from a few hours for predominantly metabolic fatigue to multiple days for muscle-damage-driven fatigue, during which biomarkers of muscle damage can be observed in blood plasma~\cite{RubioArias2019Muscle}.

Clinically, peripheral muscular fatigue is often reflected by a reduction in force-generating capacity and the appearance of delayed muscle soreness~\cite{Gussoni2023Effects}. Laboratory and clinical measurements remain the reference for quantifying these effects, including functional tests such as countermovement jumps~\cite{Springham2024Countermovement} and instrumented assessments such as bloodwork or tensiomyography~\cite{Paravlic2025Establishing}. While these approaches provide high-quality measurements, they require specialized equipment and trained personnel while not being designed for continuous monitoring during match play. Consequently, field-based tools enabling more frequent or continuous assessment have attracted increasing attention.

Among field-based approaches, wearable tracking systems such as GPS and local positioning systems (LPS) provide accurate spatio-temporal measurements that support performance-derived fatigue proxies. In particular, the \emph{acceleration--speed} (A--S) profile, often modeled as a linear relationship between instantaneous acceleration and running speed, which are essential skills in sports performance~\cite{Cardoso2023Insitu}, has been proposed as a practical indicator for tracking global fatigue over time~\cite{Morin2021Individual}. \citet{AlonsoCallejo2023Validity} evaluated the validity and reliability of the acceleration--speed (A--S) profile by comparing its derived mechanical variables with those obtained from force--velocity profiling in professional football players. In a complementary study~\cite{AlonsoCallejo2022Effect}, they investigated the influence of playing position and microcycle day on A--S parameters in elite players, reporting systematic variations across positional roles and training days. In rugby, \citet{Maviel2024Establishing} showed that significant effects can be observed on this A--S profile, and recommend practitioners to use at least $160$ minutes of actual match play to determine a reliable general A--S profile of the player. Beyond fatigue monitoring, A--S profiling has also been studied as a proxy for lower-limb mechanical capacities (\eg, hip and knee extensor strength), with potential relevance to injury prevention, including hamstring and other musculoskeletal injuries~\cite{Bramah2023Exploring}. 
In fact, returning to play after an injury may be accompanied by an A--S profile that shows a decrease in the maximal acceleration, while the maximal speed remains unchanged~\cite{Edouard2024AMusculoskeletal}. The profile is therefore also a good indicator of return to play given that hamstring injuries are prone to recurrence.
In this work, we explore whether computer vision methods can replace intrusive sensors such as GPS or LPS to derive positional information about the athletes and derive fatigue indicators in a non-intrusive way during the game.

\subsection{Computer vision in sports}

Athlete and game analysis in sports, and particularly football, has progressively shifted from manual video annotation towards automated visual understanding, guided by technological advancements in computer vision and deep learning. 
The introduction of large-scale datasets and benchmarks such as SoccerNet~\cite{Giancola2018SoccerNet} and its extensions~\cite{Deliege2021SoccerNetv2,Cioppa2022Scaling,Cioppa2022SoccerNetTracking,Giancola2022SoccerNet,Cioppa2024SoccerNet2023Challenge,Leduc2024SoccerNetDepth,Dalal2025Action, Sarkhoosh2025Beyond, Gautam2024SoccerNetEchoes}, SoccerDB~\cite{Jiang2020SoccerDB}, SoccerReplay-1988~\cite{Rao2025Towards}, AthleticPose~\cite{Suzuki2025AthleticsPose}, WorldPose~\cite{Jiang2024WorldPose}, or SoccerTrack~\cite{Scott2022SoccerTrack,Scott2025SoccerTrack-arxiv}, have supported the development of new methods and their evaluation on a broad range of tasks, including athlete segmentation~\cite{Cioppa2018ABottomUp, Cioppa2019ARTHuS}, detection\cite{Vandeghen2022SemiSupervised}, pose estimation~\cite{Ludwig2025Human}, and tracking~\cite{Maglo2022Efficient,Mansourian2023Multitask,Seweryn2025Improving,Somers2024SoccerNetGameState}, athlete re-identification~\cite{Somers2023Body,Mansourian2023Multitask,Balaji2023Jersey}, action spotting in untrimmed videos~\cite{Cioppa2020AContextaware,Cioppa2021Camera,Hong2022Spotting,Soares2022Temporally,Soares2022Action-arxiv,Giancola2023Towards,Seweryn2026Survey,Giancola2025Deep,Cabado2024Beyond,Xarles2024TDEED, Xarles2026AdaSpot-arxiv, Karki2025Pixels-arxiv}, game summarization~\cite{Gautam2022Soccer,Midoglu2024AIBased}, pass prediction and feasibility analysis~\cite{ArbuesSanguesa2020Using,Honda2022Pass}, foul recognition~\cite{Held2023VARS,Held2025Towards,Held2024XVARS,Held2025Enhancing,Fang2025Foul}, depth estimation~\cite{Leduc2024SoccerNetDepth}, dense captioning for broadcast commentary~\cite{Mkhallati2023SoccerNetCaption,Andrews2024AiCommentator}, 3D understanding~\cite{GutierrezPerez2025SoccerNetv3D}, and visual question answering leveraging large Video Language Models (VLMs)~\cite{Yang2026SoccerMaster}.

Recent advances in multiple-object tracking and player identification in broadcast videos~\cite{Cioppa2022SoccerNetTracking,Maglo2023Individual,Balaji2023Jersey}, together with camera calibration methods exploiting pitch geometry~\cite{Theiner2021Extraction,Magera2024AUniversal,GutierrezPerez2024NoBells,GutierrezPerez2026PnLCalib,Magera2025CanGeometry,Magera2025BroadTrack}, enable player detections to be projected into a metric coordinate system. This makes it possible to reconstruct spatio-temporal trajectories and derive kinematic quantities such as speed and positional distributions over extended match sequences~\cite{Rustebakke2025Extracting}. The Game State Reconstruction (GSR) benchmark~\cite{Somers2024SoccerNetGameState} integrates these components into a unified framework, with Broadcast2Pitch~\cite{Oo2026Broadcast2Pitch} and SoccerMaster~\cite{Yang2026SoccerMaster} currently representing state-of-the-art approaches to this task.
In this work, we build upon these GSR pipelines to extract metric player trajectories and introduce a speed and acceleration estimation procedure designed to construct acceleration-speed (A--S) profiles from broadcast video. These profiles are then analyzed as fatigue-related kinematic indicators.

\section{Methodology}
\label{sec:method}

\paragraph{Problem statement.}
Given the output of a Game State Reconstruction (GSR) method, our goal is to build acceleration-speed (A–S) profiles to model the physical load and fatigue of athletes over time. Specifically, for a given video clip of a football game, a GSR pipeline provides the localization and identity of all athletes on the pitch at each frame. This global task typically involves several sub-problems, including athlete detection, camera calibration and homography estimation, multi-object tracking, and jersey number recognition. As a result, each athlete can be identified and localized in 2D field coordinates across time.

More formally, let a video clip be composed of $T$ frames sampled at frame rate $f$, with timestamps $t_k = k / f$, with $k = \{1, \dots, T$\}.  
For each athlete $i \in \{1, \dots, N\}$, the GSR method outputs temporal 2D positions on the field:
\begin{equation}
\mathbf{p}_i(t_k) = (x_i(t_k), y_i(t_k)) \in \mathbb{R}^2,
\end{equation}
expressed in metric field coordinates.  
The objective is then to transform these athlete trajectories into the physical signals of \emph{speed} and \emph{acceleration}, enabling the building of A--S profiles and modeling of the athlete's physical load.

\begin{figure*}[t]
    \centering
    \scriptsize
    \vspace{-1.5em}
    \begin{minipage}[t]{0.485\textwidth}
        \centering
        \includegraphics[width=\linewidth,trim={0cm 0cm 0cm 1.75cm},clip]{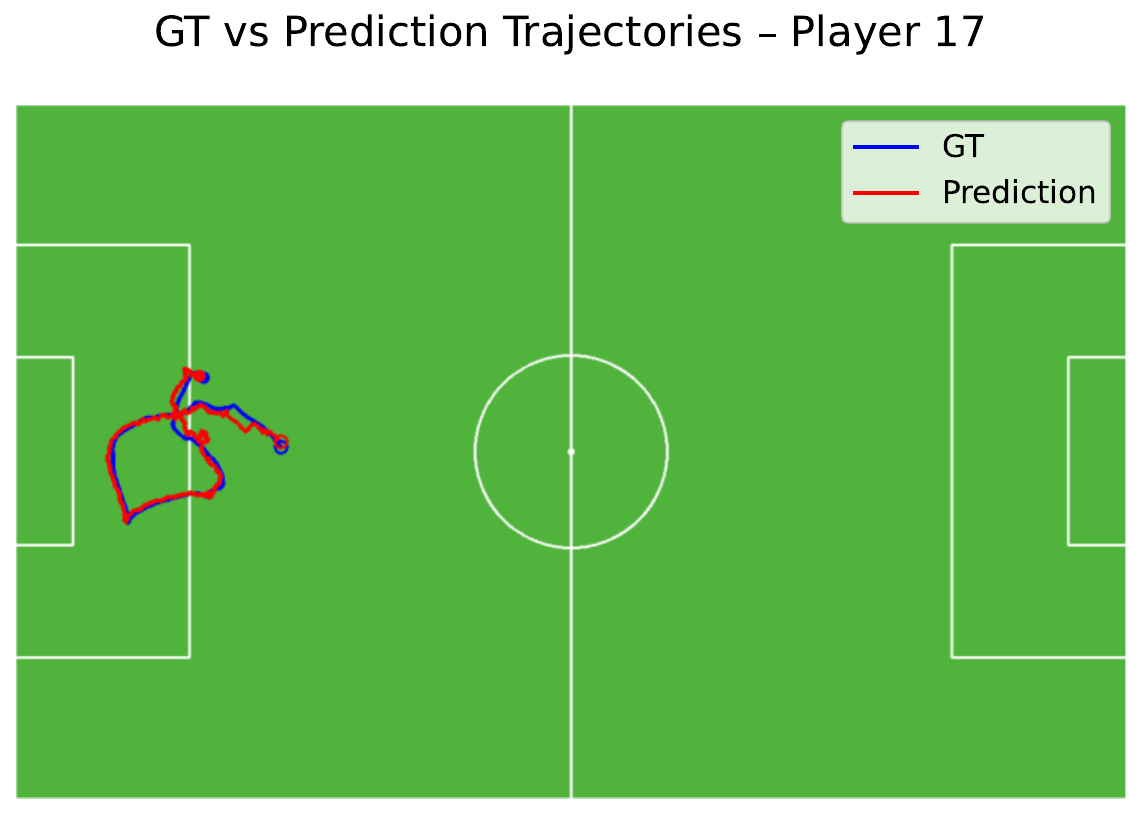}
    \end{minipage}%
    \hspace{0.02\textwidth} %
    \begin{minipage}[t]{0.485\textwidth}
        \centering
        \includegraphics[width=\linewidth,trim={0cm 0cm 0cm 1.75cm},clip]{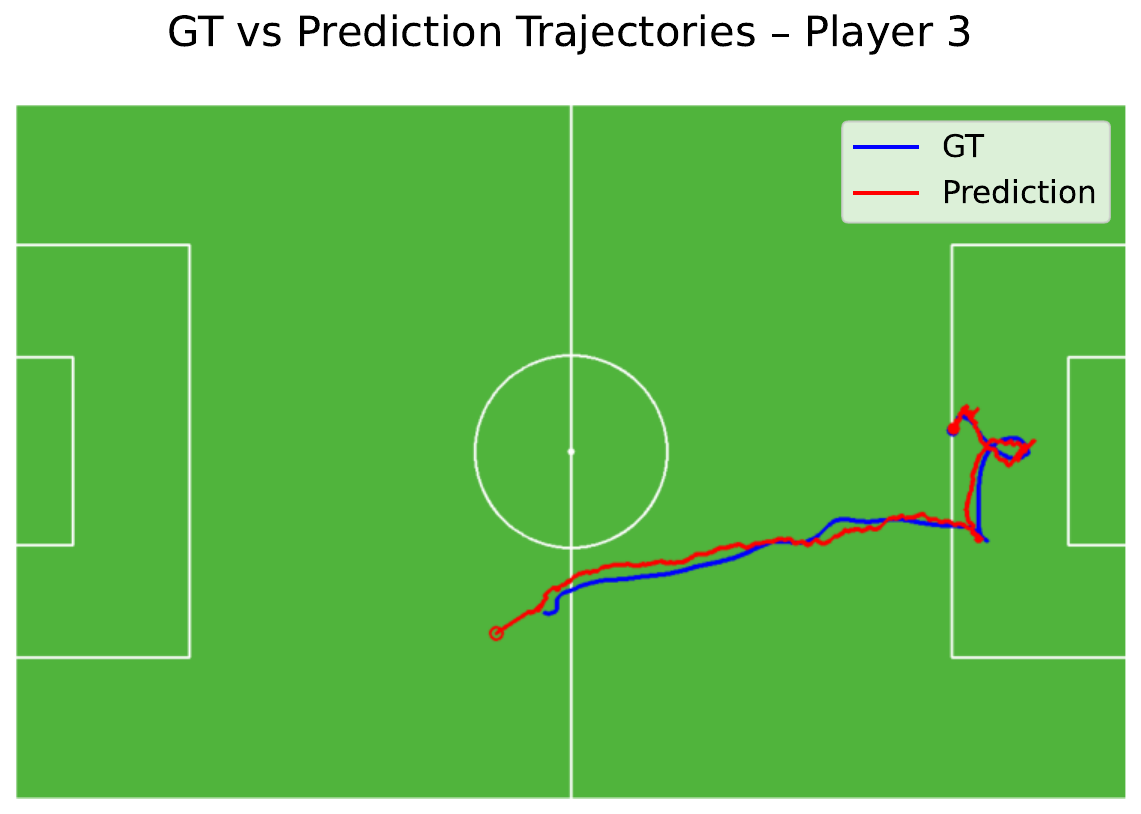}
    \end{minipage}
    \caption{\textbf{Qualitative examples.} Qualitative examples of predicted athlete trajectories (red) and their respective ground-truth trajectories (blue) on the SoccerNet Game State Reconstruction (GSR) test set. (a) Depicts the trajectory of player $17$ in the SNGS-200 sequence, and (b) portrays the one of player $3$ in the SNGS-199 sequence. As observed, although the predicted trajectories exhibit minor localization noise, they closely follow the ground-truth trajectories.}
    \label{fig:trajectory_pred_examples}
\end{figure*}

\subsection{Extracting athlete trajectories}

Given per-frame detections, a game state reconstruction method predicts for each athlete $i$ a discrete trajectory:
\begin{equation}
\mathcal{T}_i = \{\mathbf{p}_i(t_k)\}_{k=1}^{\upperBoundT},\end{equation}
where some $\mathbf{p}_i(t_k)$ might be missing due to possible occlusions, missed detections, or the athlete leaving the field of view. As GSR predictions may contain detection errors or localization noise, we apply the following post-processing to the athlete's trajectory:

\begin{enumerate}
    \item \textbf{Short-gap interpolation.}  
    We linearly interpolate missing positions for short temporal gaps of length at most $g_{\max}=3$ consecutive frames. At a standard $25 fps$, this corresponds to 0.12 $s$. Longer gaps are ignored, as the athlete may have simply exited the field of view.

    \item \textbf{Temporal smoothing.}  
    To reduce high-frequency localization noise, we apply a temporal smoothing operator $\mathcal{S}$. We consider two alternatives:  
    (i) a Kalman filter~\cite{Kalman1960ANew}, and  
    (ii) a Savitzky–Golay filter~\cite{Savitzky1964Smoothing}.  
\end{enumerate}
The resulting post-processed trajectory is 
\begin{equation}
\tilde{\mathcal{T}}_i = \{\tilde{\mathbf{p}}_i(t_k)\}_{k=1}^{\upperBoundT},
\end{equation}
which serves as the basis for the calculation of speed and acceleration signals.

\subsection{Computation of speed and acceleration}

Given a processed trajectory 
\(
\tilde{\mathcal{\upperBoundT}}
\), 
we compute speed as the derivative of position and acceleration as the derivative of speed. Direct frame-to-frame differences at typical frame rates (\eg, $f = 25$ Hz, $\Delta t = 40$ ms) produce high-frequency noise signals as the difference operation acts as a high-pass filter. This effect is particularly amplified when estimating the acceleration due to the second-order derivative. To mitigate this effect, we estimate motion over a broader temporal window of $0.5$-$2$ s as explained hereafter.

Let $l_n$ denote the set of indexes of the temporal neighborhood around $k$ that is used for differentiation. For each valid index $k$, we define 
\begin{equation}
k^- = \max(k-l_n, 1), 
\qquad
k^+ = \min(k+l_n, \upperBoundT),
\end{equation}
and compute the velocity vector as:
\begin{equation}
\mathbf{v}_i(t_k) = \frac{\tilde{\mathbf{p}}_i(t_{k^+}) - \tilde{\mathbf{p}}_i(t_{k^-})}{t_{k^+} - t_{k^-}}.
\end{equation}
Subsequently, the acceleration is computed analogously from the velocity signal:
\begin{equation}
\mathbf{a}_i(t_k) = \frac{\mathbf{v}_i(t_{k^+}) - \mathbf{v}_i(t_{k^-})}{t_{k^+} - t_{k^-}}.
\end{equation}
The parameter $l_n$ controls the temporal differentiation scale, with larger values yielding smoother signals, while excessively large values may over-attenuate the signal, reducing local variations that are informative for fatigue profiles. 

Finally, speed and acceleration signals are post-processed in the following way: (1) speeds above 15 m/s are removed to discard unrealistic jumps in the trajectories, and (2) acceleration signals are smoothed with a central moving average over $20$ values, following the recommendations of~\citet{Pons2021Integrating}, to reduce high-frequency noise.

\subsection{Building an A--S profile}
The acceleration–speed (A--S) profile describes the linear relationship between an athlete's running speed and the corresponding maximal acceleration generated over a defined monitoring period~\cite{Morin2021Individual}. The resulting model is characterized by three variables:
(1) the theoretical maximal acceleration $A_0$, which is the interception point with the y-axis. It represents the maximal acceleration capacity at low speeds, (2) the theoretical maximal running speed $S_0$, \ie the interception point with the x-axis, representing the speed for which acceleration is $0$, and (3) the A--S slope, which reflects the general balance between acceleration and speed capacities, \ie how acceleration decreases as athlete speed increases.
The interpretation of the A--S profile is straightforward. High $A_0$ reflects a greater capability to produce short, explosive actions, while high $S_0$ is correlated with greater top-speed capacity~\cite{JimenezReyes2020Seasonal}. In addition, steeper A--S slopes indicate a more acceleration-oriented profile, while flatter slopes represent velocity-oriented profiles. 

The evolution of the A--S profile during a game can serve as an objective indicator of fatigue, as the athlete may progressively lose the ability to generate high accelerations. Given the athlete-specific speed and acceleration signals, we construct individual A--S profiles following the methodology of \citet{Morin2021Individual}. Acceleration is plotted as a function of speed, and appropriate samples are selected for A--S modeling. We retain running speeds between $3$~m/s and the athlete’s maximal speed. For each $0.2$~m/s speed bin, we select the two highest acceleration values. A linear regression is first fitted to the selected points, and outliers are removed using a $95$\% confidence interval. A final linear regression is then performed on the filtered samples.

\section{Experiments}

\begin{table*}[t]
\centering
\caption{\textbf{Ablation study on the effect of the filtering method on predicted athlete speed and acceleration.} Results are reported on the test set of the SoccerNet Game State Reconstruction (GSR) benchmark. We show MAE, RMSE, Pearson correlation ($r$), and max-lag $r$ at $25$~fps. Arrows indicate whether higher values correspond to better (↑) or worse (↓) performance.}
\label{tab:ablation_filter_method}
\begin{tabular*}{\linewidth}{@{\extracolsep{\fill}}cccccccccc}
\toprule
\textbf{$l_n$} & \textbf{Filter} & \multicolumn{4}{c}{\textbf{Speed}} & \multicolumn{4}{c}{\textbf{Acceleration}} \\
\cmidrule(lr){3-6} \cmidrule(lr){7-10}
 &  & \textbf{MAE ↓} & \textbf{RMSE ↓} & $\textbf{\textit{r} ↑}$ & \textbf{max-lag $r$ ↑} & \textbf{MAE ↓} & \textbf{RMSE ↓} & $\textbf{\textit{r} ↑}$ & \textbf{max-lag $r$ ↑} \\
\cmidrule(lr){3-3} \cmidrule(lr){4-4} \cmidrule(lr){5-5} \cmidrule(lr){6-6} 
\cmidrule(lr){7-7} \cmidrule(lr){8-8} \cmidrule(lr){9-9} \cmidrule(lr){10-10}

  & None   & 0.776 & 1.364 & 0.732 & 0.767 & 1.857 & 2.548 & 0.301 & 0.370 \\
5  & Kalman & \textbf{0.761} & \textbf{1.350} & \textbf{0.739} & \textbf{0.773} & 1.788 & 2.470 & 0.320 & 0.390 \\
  & Savgol & 0.768 & 1.389 & 0.724 & 0.761 & \textbf{1.758} & \textbf{2.439} & \textbf{0.327} & \textbf{0.398} \\
\midrule

  & None   & 0.610 & 1.196 & 0.814 & 0.841 & 0.934 & 1.493 & 0.523 & 0.580 \\
10 & Kalman & \textbf{0.607} & \textbf{1.192} & \textbf{0.816} & \textbf{0.842} & 0.926 & \textbf{1.486} & \textbf{0.531} & \textbf{0.589} \\
  & Savgol & 0.615 & 1.226 & 0.799 & 0.826 & \textbf{0.924} & 1.491 & 0.522 & 0.581 \\
\midrule

  & None   & \textbf{0.485} & 1.103 & 0.878 & 0.895 & 0.452 & 0.938 & 0.712 & 0.745 \\
20 & Kalman & \textbf{0.485} & 1.103 & \textbf{0.879} & \textbf{0.896} & \textbf{0.451} & 0.932 & \textbf{0.716} & \textbf{0.749} \\
  & Savgol & 0.486 & \textbf{1.101} & 0.869 & 0.887 & \textbf{0.451} & \textbf{0.931} & 0.705 & 0.740 \\
\bottomrule
\end{tabular*}%
\end{table*}

\begin{table*}[t]
\centering
\caption{\textbf{Ablation study on the effect of the local neighborhood $l_n$ on predicted athlete speed and acceleration.} Results are reported on the test set of the SoccerNet GSR benchmark. We report MAE, RMSE, Pearson correlation ($r$), and max-lag $r$ at 25~fps. Results are shown as mean $\pm$ standard deviation. Arrows indicate whether higher values correspond to better (↑) or worse (↓) performance.}
\label{tab:ablation_n_neighbors}
\resizebox{\linewidth}{!}{%
\begin{tabular}{cccccccccc} 
\toprule
$l_n$ & \textbf{Temporal} & \multicolumn{4}{c}{\textbf{Speed}} & \multicolumn{4}{c}{\textbf{Acceleration}} \\
\cmidrule(lr){3-6} \cmidrule(lr){7-10}
 & \textbf{window (s)} & \textbf{MAE ↓} & \textbf{RMSE ↓} & $\textbf{\textit{r} ↑}$ & \textbf{max-lag $r$ ↑} & \textbf{MAE ↓} & \textbf{RMSE ↓} & $\textbf{\textit{r} ↑}$ & \textbf{max-lag $r$ ↑} \\
\cmidrule(lr){3-3} \cmidrule(lr){4-4} \cmidrule(lr){5-5} \cmidrule(lr){6-6} 
\cmidrule(lr){7-7} \cmidrule(lr){8-8} \cmidrule(lr){9-9} \cmidrule(lr){10-10}

2  & 0.20 & 1.067 $\pm$ 1.361 & 1.729 $\pm$ 1.361 & 0.616 $\pm$ 0.380 & 0.652 $\pm$ 0.373 
   & 3.442 $\pm$ 2.358 & 4.172 $\pm$ 2.358 & 0.094 $\pm$ 0.150 & 0.195 $\pm$ 0.131 \\

5  & 0.44 & 0.761 $\pm$ 1.115 & 1.350 $\pm$ 1.115 & 0.739 $\pm$ 0.495 & 0.773 $\pm$ 0.481 
   & 1.788 $\pm$ 1.704 & 2.470 $\pm$ 1.704 & 0.320 $\pm$ 0.233 & 0.390 $\pm$ 0.219 \\

10 & 0.80 & 0.607 $\pm$ 1.026 & 1.192 $\pm$ 1.026 & 0.816 $\pm$ 0.613 & 0.842 $\pm$ 0.581 
   & 0.926 $\pm$ 1.163 & 1.486 $\pm$ 1.163 & 0.531 $\pm$ 0.389 & 0.589 $\pm$ 0.384 \\

15 & 1.24 & 0.529 $\pm$ 1.000 & 1.131 $\pm$ 1.000 & 0.858 $\pm$ 0.696 & 0.876 $\pm$ 0.665 
   & 0.601 $\pm$ 0.920 & 1.099 $\pm$ 0.920 & 0.646 $\pm$ 0.511 & 0.687 $\pm$ 0.501 \\

20 & 1.64 & 0.485 $\pm$ 0.991 & 1.103 $\pm$ 0.991 & 0.879 $\pm$ 0.771 & 0.896 $\pm$ 0.739 
   & 0.451 $\pm$ 0.815 & 0.932 $\pm$ 0.815 & 0.716 $\pm$ 0.597 & 0.749 $\pm$ 0.586 \\

\bottomrule
\end{tabular}%
}
\end{table*}

\begin{figure*}[t]
    \centering
    \scriptsize
    \vspace{-1.5em}

    \begin{subfigure}[t]{0.31\textwidth}
        \centering
        \caption{Reliable speed, sequence SNGS-188}
        \includegraphics[width=\linewidth,trim={0 0 0 0cm},clip]{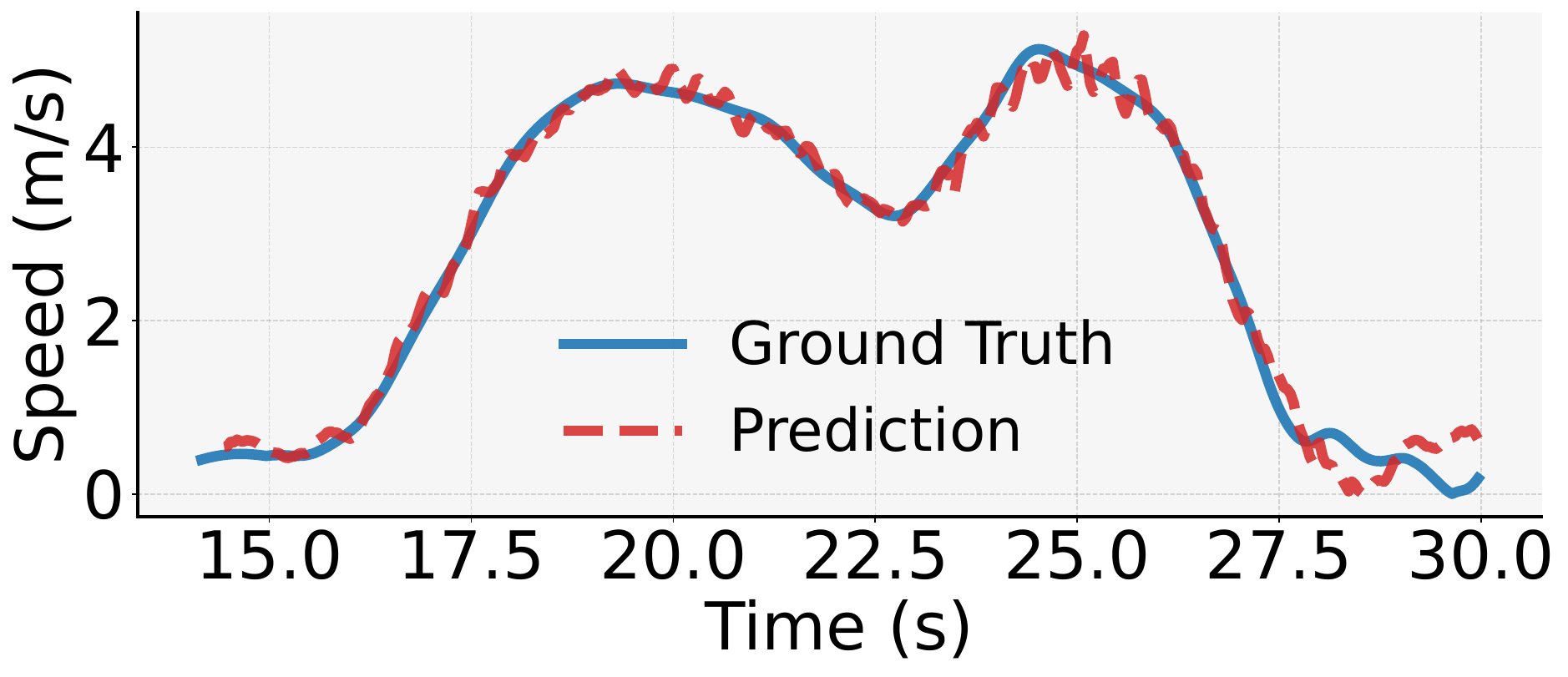}
        \label{fig:reliable_speed_1}
    \end{subfigure}
    \hspace{0.02\textwidth}
    \begin{subfigure}[t]{0.31\textwidth}
        \centering
        \caption{Reliable speed, sequence SNGS-119}
        \includegraphics[width=\linewidth,trim={0 0 0 0cm},clip]{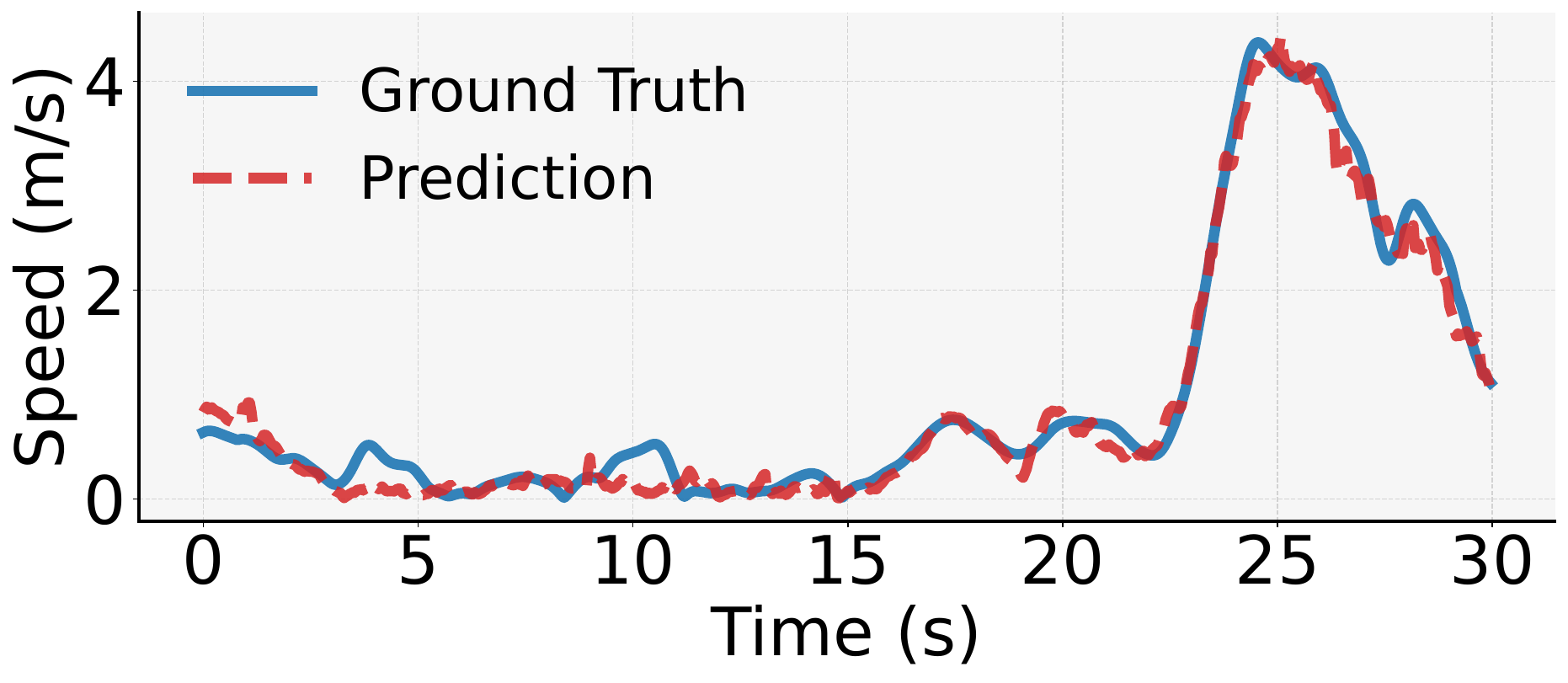}
        \label{fig:reliable_speed_2}
    \end{subfigure}
    \hspace{0.02\textwidth}
    \begin{subfigure}[t]{0.31\textwidth}
        \centering
        \caption{Unreliable speed, sequence SNGS-148}
        \includegraphics[width=\linewidth,trim={0 0 0 0cm},clip]{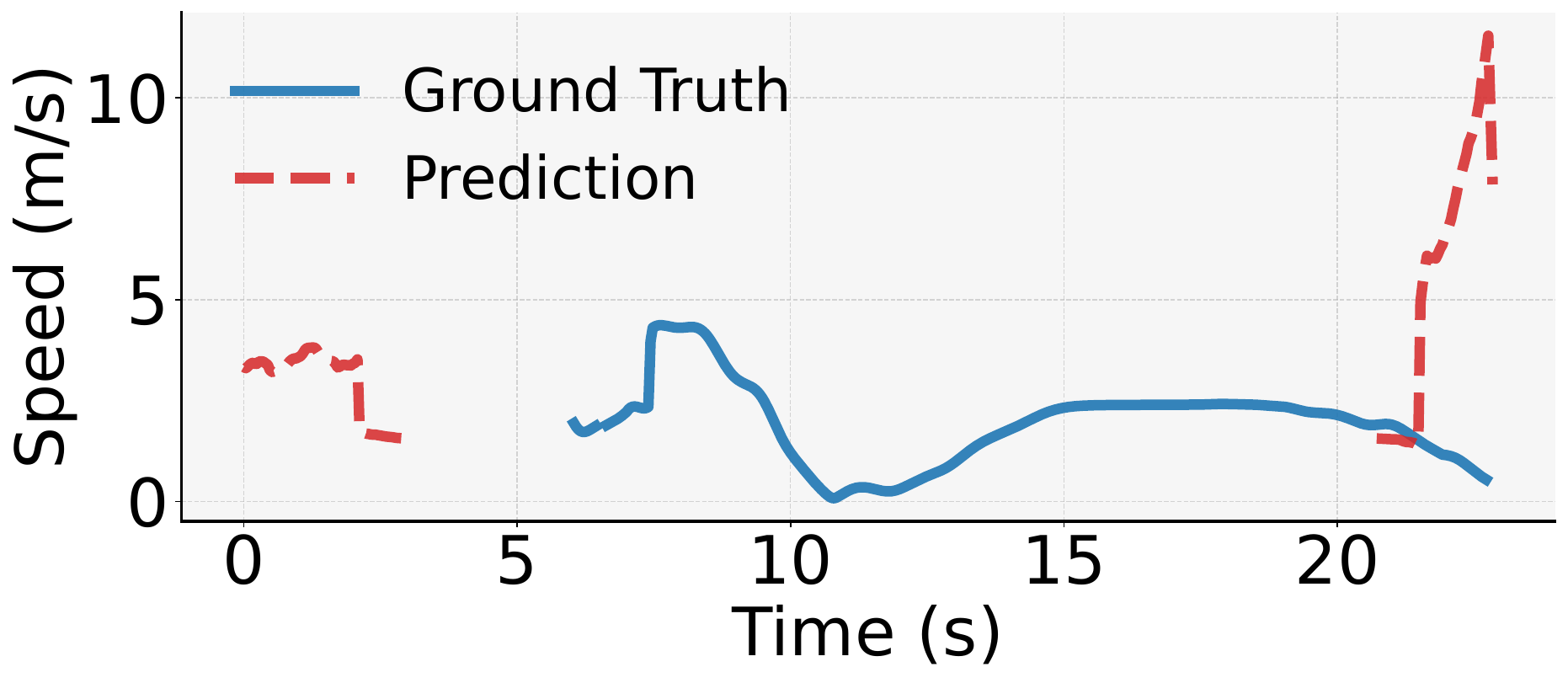}
        \label{fig:unreliable_speed_1}
    \end{subfigure}
    \hspace{0.02\textwidth}
    
    \vspace{0.01em}
    
    \begin{subfigure}[t]{0.31\textwidth}
        \centering
        \caption{Reliable acceleration, sequence SNGS-188}
        \includegraphics[width=\linewidth,trim={0 0 0 0cm},clip]{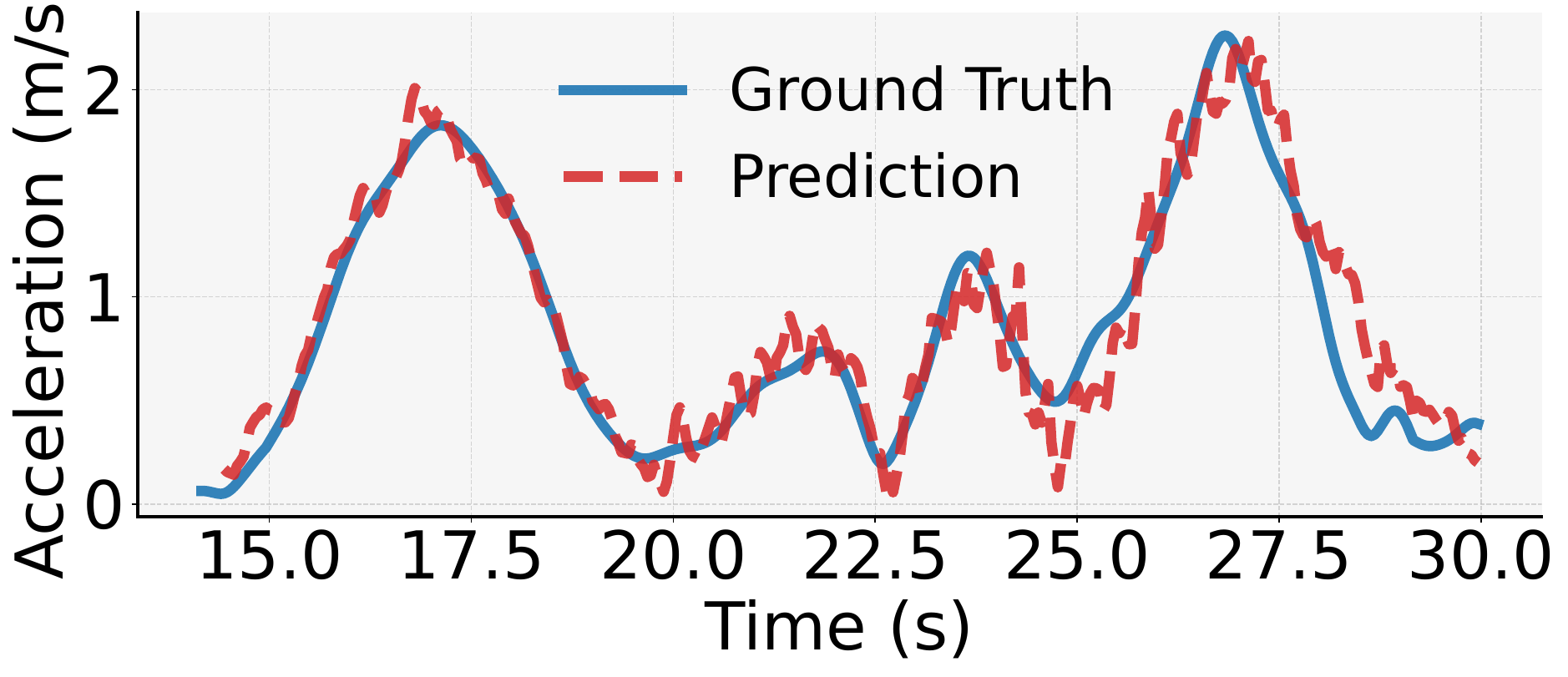}
        \label{fig:reliable_acc_1}
    \end{subfigure}
    \hspace{0.02\textwidth}
    \begin{subfigure}[t]{0.31\textwidth}
        \centering
        \caption{Reliable acceleration, sequence SNGS-119}
        \includegraphics[width=\linewidth,trim={0 0 0 0cm},clip]{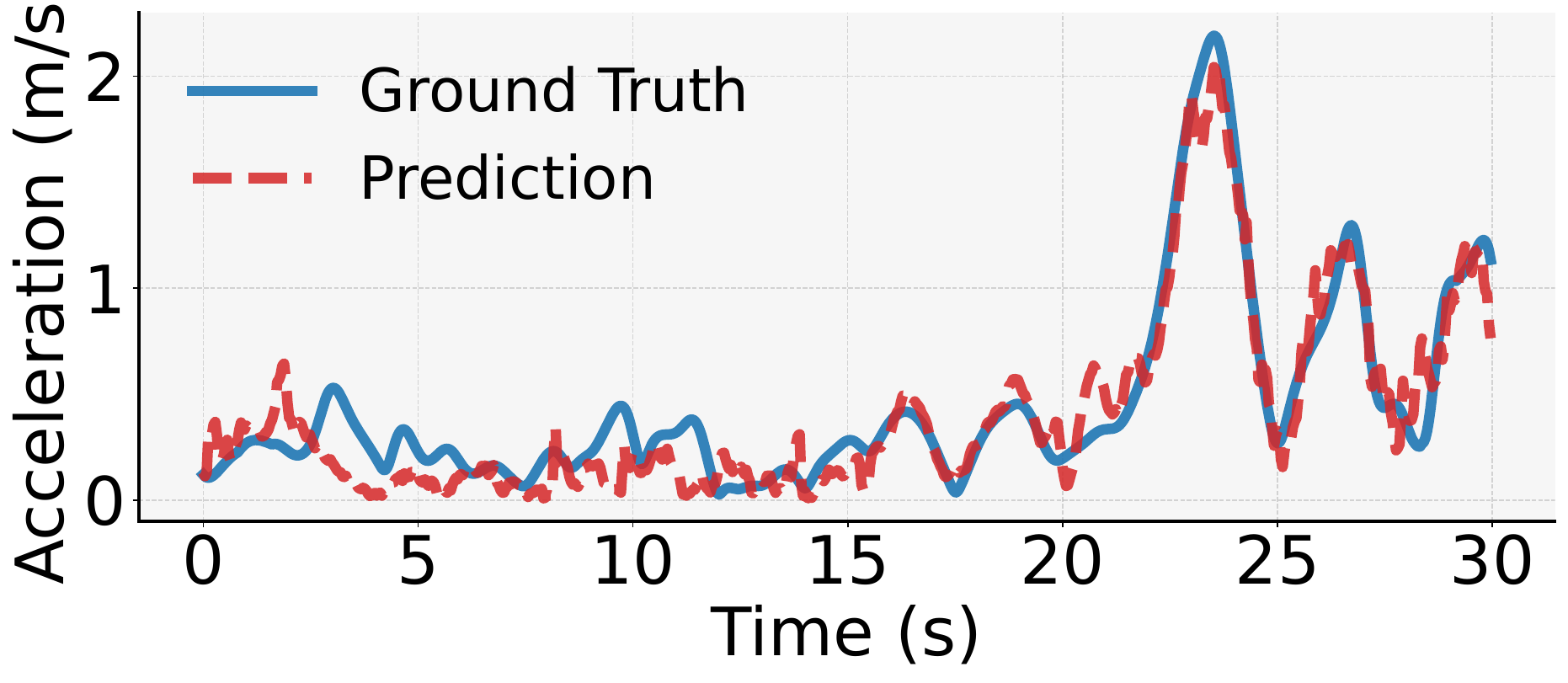}
        \label{fig:reliable_acc_2}
    \end{subfigure}
    \hspace{0.02\textwidth}
    \begin{subfigure}[t]{0.31\textwidth}
        \centering
        \caption{Unreliable acceleration, sequence SNGS-148}
        \includegraphics[width=\linewidth,trim={0 0 0 0cm},clip]{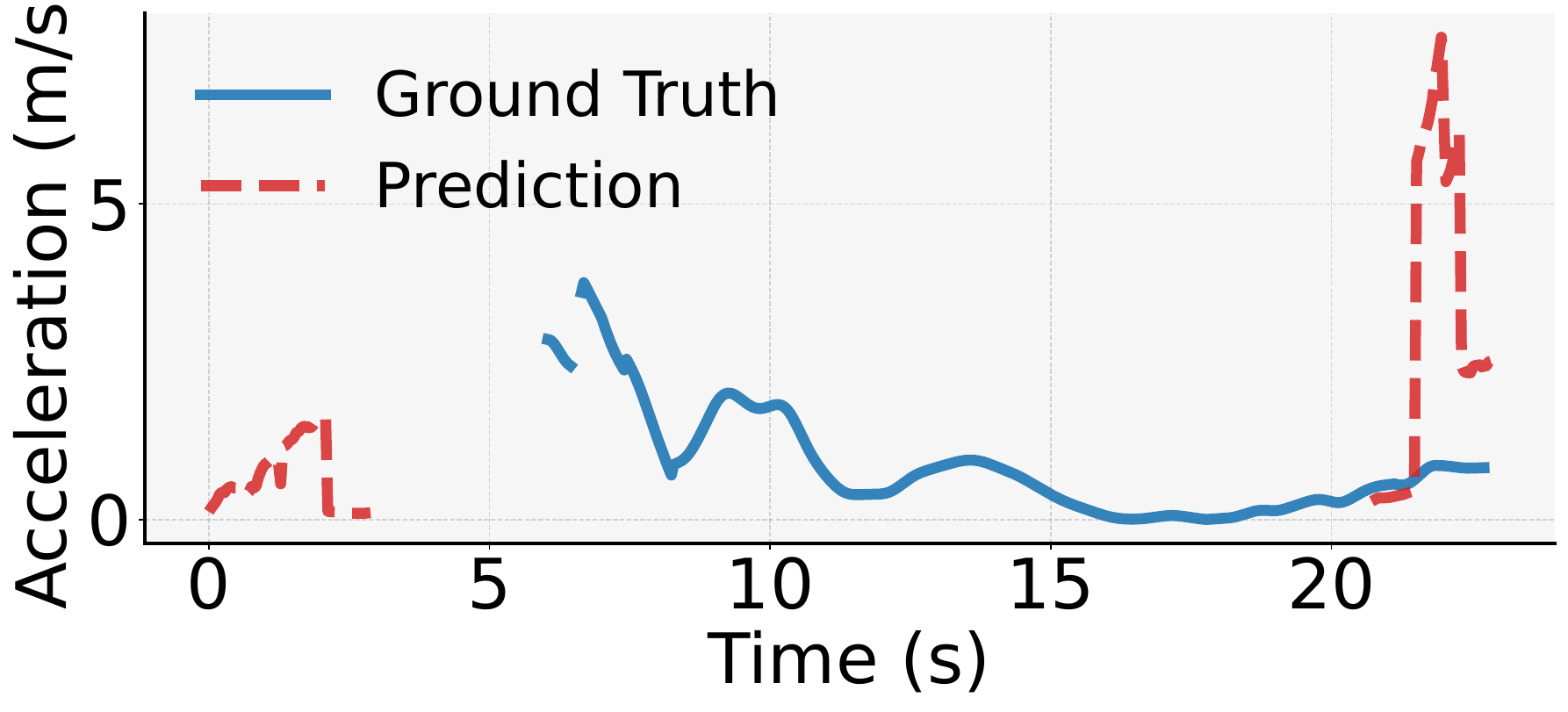}
        \label{fig:unreliable_acc_1}
    \end{subfigure}

    \caption{
    \textbf{Qualitative examples of reliable and non-reliable speed and acceleration predictions.}
    The first two columns depict two athletes with high reliability scores showing strong agreement between predictions and ground truth for both speed and acceleration.
    The last column shows a low-reliability case where the athlete was not detected for most of the sequence, leading to large error.
    }
    \label{fig:reliable_vs_unreliable}
\end{figure*}

\mysection{Data.} %
Experiments are conducted on the SoccerNet GSR test set~\cite{Somers2024SoccerNetGameState}, which provides 2D field positions for all athletes across $49$ video sequences, including $773$ athletes. From these positions, we derive speed and acceleration ground-truth signals as described in \cref{sec:method}, without any filtering or smoothing, and using the $l_n$ temporal window. Moreover, we provide an additional experiment on a full $45$-minute half-game from the Swiss Super League to build the A--S profile of athletes over a longer temporal window. 

\mysection{Trajectory estimation.} We process sequences from the SoccerNet GSR test set using the current state-of-the-art method Broadcast2Pitch~\cite{Oo2026Broadcast2Pitch} and apply our methodology to the resulting predicted athlete trajectories.
Out of $746$ predicted athletes, we match $578$ to ground-truth trajectories. 
\Cref{fig:trajectory_pred_examples} presents two representative predicted trajectories from the GSR SoccerNet test set. In addition to the predictions, we also overlay the ground-truth trajectories to enable direct quantitative comparison. Although the predicted trajectories exhibit minor localization noise, they closely follow the ground-truth trajectories.

\mysection{Evaluation metrics.} We evaluate the predicted athlete speed and acceleration across all video clips using both amplitude- and correlation-based metrics. To measure per-sample deviations, we report the mean absolute error (MAE) and the root mean squared error (RMSE), which are commonly used in signal processing and to compare camera-based and GPS-based measurements~\cite{Pons2021Integrating}. While accurate amplitudes are important, capturing the temporal dynamics of motion is often more critical, as relative changes enable the detection of direction changes and sprint events. To this end, we also report the Pearson correlation \(r\) and the max-lag \(r\), which quantify how well the predicted signals align with the temporal structure of the ground truth. To summarize correlations across multiple athletes, we apply the Fisher $z$-transform~\cite{Fisher1915Frequency} before averaging:
\begin{equation}
    z = \frac{1}{2} \ln\frac{1+r}{1-r}, \quad r_\text{avg} = \frac{e^{2 \bar{z}}-1}{e^{2 \bar{z}}+1}.
\end{equation}
This transformation avoids bias in the mean correlation due to the nonlinear distribution of $r$ values, providing a statistically sound estimate of overall temporal correlation.

\mysection{Results on speed and acceleration estimation.}
A quantitative evaluation of the predicted speed and acceleration signals is conducted under different design choices. \Cref{tab:ablation_filter_method} compares the impact of the temporal filtering step: no filtering, a constant-velocity Kalman filter, and a Savitzky--Golay filter with window size of $9$ and a polynomial order of $2$. Results indicate that both filtering strategies improve performance, particularly for smaller values of $l_n$, which produce noisier signals. However, as $l_n$ increases, the difference between results decreases. The Kalman filter provides better $r$ and max-lag $r$, \ie better temporal consistency with respect to the ground truth, which is essential for reliable fatigue modeling over extended time periods. Consequently, we adopt the Kalman filter in the remainder of our experiments.

Moreover, we provide the speed and acceleration quantitative results for different values of $l_n$ in \cref{tab:ablation_n_neighbors}. As expected, increasing $l_n$ improves MAE, RMSE, $r$, and max-lag $r$, as larger local neighborhoods reduce noise. For $l_n=2$, \ie a span of\,$0.2$ seconds, the signal error and temporal inconsistency are considerably large. As can be seen, performance steadily improves with larger $l_n$. Hence, we select a maximum value of $l_n=20$, corresponding to a temporal span of $1.64$\,s. Larger values are avoided to prevent excessive smoothing and loss of informative local motion dynamics. 

\mysection{Error analysis on spatio-temporal predictions.} The obtained quantitative speed and acceleration results in \cref{tab:ablation_n_neighbors} indicate a large variability among athletes. For better interpretability, we conduct reliability analysis of the predictions to address to what extent one can trust GSR spatio-temporal predictions. Thus, we define a prediction reliability measure as a combination of speed scores
\begin{equation}
    R = R_c \cdot R_\text{MAE} \cdot R_d,
\end{equation}
where $R_c$ is the Pearson correlation of predicted versus ground-truth speed, 
$R_\text{MAE} = 1/(1 + \text{MAE}_\text{speed})$ penalizes large amplitude errors, 
and $R_d =  N_\text{frames}/N_\text{total\_frames}$ weights the score by the fraction of frames in which the athlete is detected by the GSR method. 
This reliability score allows us to identify the 30\% most reliable and 30\% least reliable athlete spatio-temporal detections (173 athlete trajectories in each case), which we analyze separately in \cref{tab:reliability_analysis_top_bottom}.

As observed, the top $30$\% most reliable detections achieve high mean scores with low variance, particularly for speed measurements. In contrast, the bottom $30$\% of detections exhibit substantially lower performance and higher variability, as expected. Moreover, the temporal consistency of the most reliable detections, quantified by Pearson $r$, remains robust for both speed and acceleration. This is critical for applications such as fatigue assessment, where accurately capturing sprints and rapid accelerations is essential.

These results highlight that current state-of-the-art GSR methods produce spatio-temporal athlete detections with considerable variability in performance. Consequently, extracting speed and acceleration signals from the most reliable detections could enable robust acceleration–speed (A--S) profile construction from monocular video. Nevertheless, prediction errors are frequent in challenging scenarios, such as occlusions or cluttered areas. In these cases, predicted trajectories are insufficiently reliable for A--S profile extraction, suggesting that future work should focus on identifying and mitigating unreliable detections. 
\Cref{fig:reliable_vs_unreliable} illustrates both reliable and unreliable examples. In the unreliable case, the athlete is missed for most of the sequence, resulting in large speed and acceleration errors. In contrast, examples from the top $30$\% most reliable detections exhibit strong temporal correlation with ground truth signals. Although acceleration remains noisier than speed, it still shows meaningful agreement, consistent with the quantitative results.

\begin{table}[t]
\centering
\caption{\textbf{Reliability analysis of athlete motion dynamics prediction.} on the test set of the SoccerNet Game State Reconstruction (GSR) benchmark. Values are reported as mean $\pm$ standard deviation. Max-lag $r$ is not defined for some low-reliability athletes.}
\label{tab:reliability_analysis_top_bottom}
\resizebox{\linewidth}{!}{%
\begin{tabular}{lcc}
\toprule
\textbf{Metric} & \textbf{\shortstack{Top 30\% \\ Reliable athletes}} & \textbf{\shortstack{Bottom 30\% \\ Reliable athletes}} \\
\midrule
Avg. visible frames & 537 (21.48\,s) & 351 (14.04\,s)\\ 
\midrule
\multicolumn{3}{c}{\textbf{Speed}} \\
\midrule
Pearson $r$ & 0.975 $\pm$ 0.021 & 0.489 $\pm$ 0.327 \\
MAE        & 0.234 $\pm$ 0.055 & 1.078 $\pm$ 0.751 \\
RMSE       & 0.132 $\pm$ 0.073 & 4.545 $\pm$ 5.554 \\
\midrule
\multicolumn{3}{c}{\textbf{Acceleration}} \\
\midrule
Pearson $r$ & 0.896 $\pm$ 0.137 & 0.375 $\pm$ 0.309 \\
MAE        & 0.263 $\pm$ 0.099 & 0.881 $\pm$ 0.600 \\
RMSE       & 0.270 $\pm$ 0.669 & 2.714 $\pm$ 3.561 \\
\bottomrule
\end{tabular}
}
\end{table}

\begin{figure*}[t]
    \centering
    \scriptsize
    \begin{subfigure}[t]{0.485\textwidth}
        \centering
        \includegraphics[width=\linewidth,trim={0 0 0 2.15cm},clip]{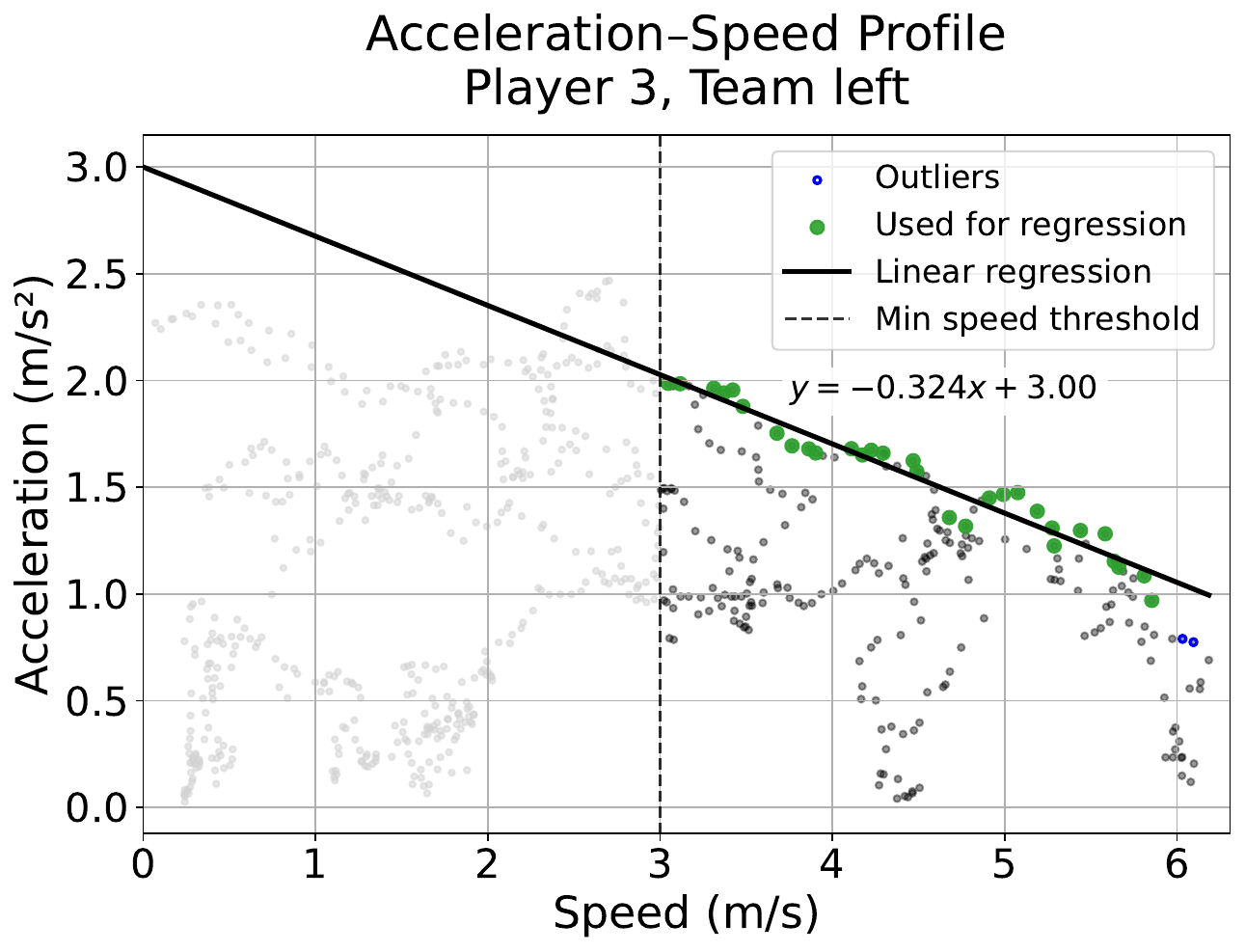}
        \caption{A--S profile derived from predicted athlete speed and acceleration.}
        \label{fig:unreliable_speed_2}
    \end{subfigure}
    \hspace{0.02\textwidth}
    \begin{subfigure}[t]{0.485\textwidth}
        \centering
        \includegraphics[width=\linewidth,trim={0 0 0 2.15cm},clip]{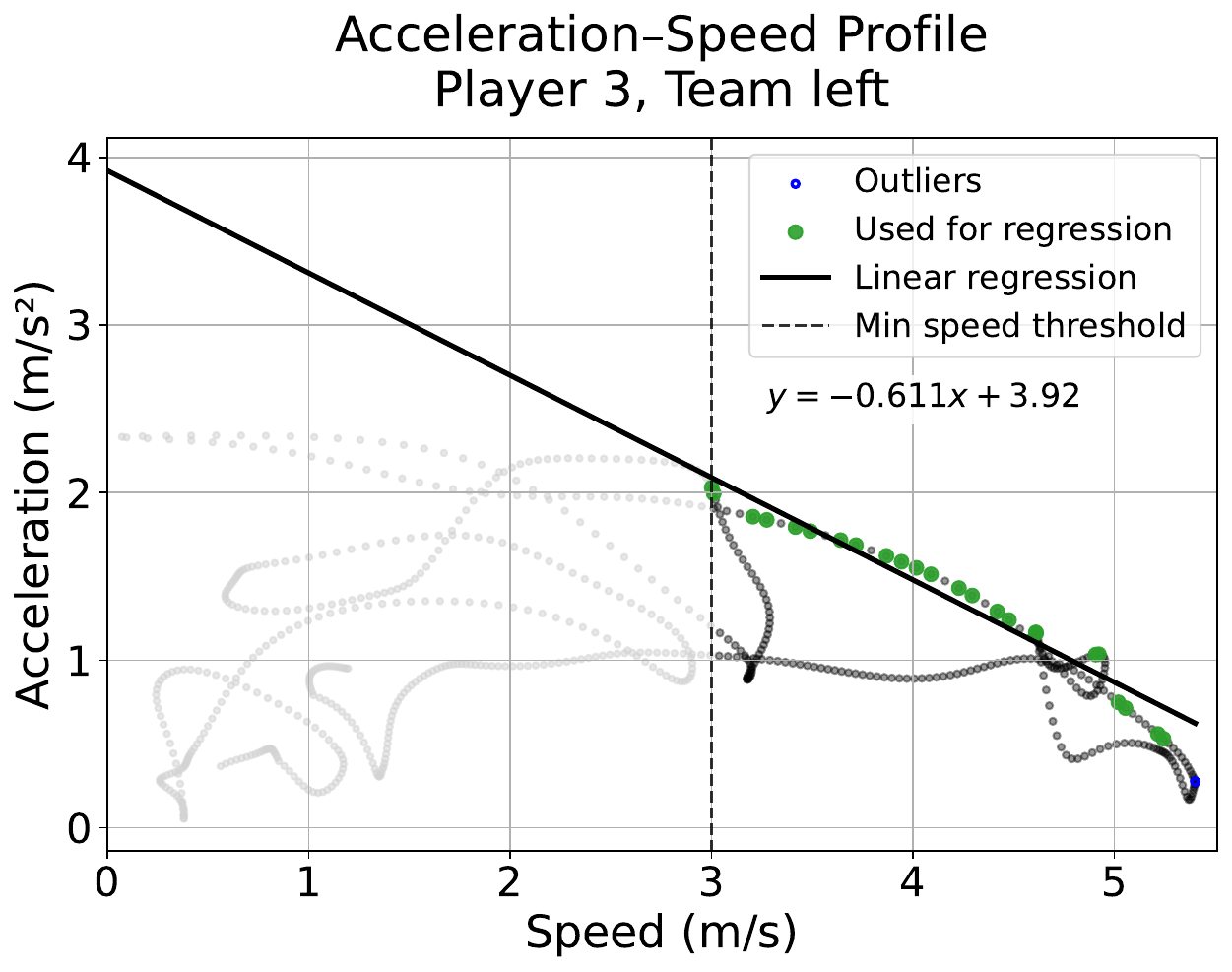}

        \caption{Ground truth A--S profile derived from reference athlete trajectory.}
        \label{fig:unreliable_acc_2}
    \end{subfigure}
    \caption{\textbf{Comparison between predicted and ground truth A--S profiles} for sequence SNGS-199, team left, jersey 3. As can be seen, predicted A--S curve follows the ground truth signal, despite clear fluctuations due to noise and accumulated errors. Green points are the ones selected to regress the linear equation.}
    \label{fig:AP_profile_SNGS-199}
\end{figure*}

\begin{figure*}[t]
    \centering
    \scriptsize
    \begin{subfigure}[t]{0.485\textwidth}
        \centering
        \includegraphics[width=\linewidth,trim={0 0 0 0cm},clip]{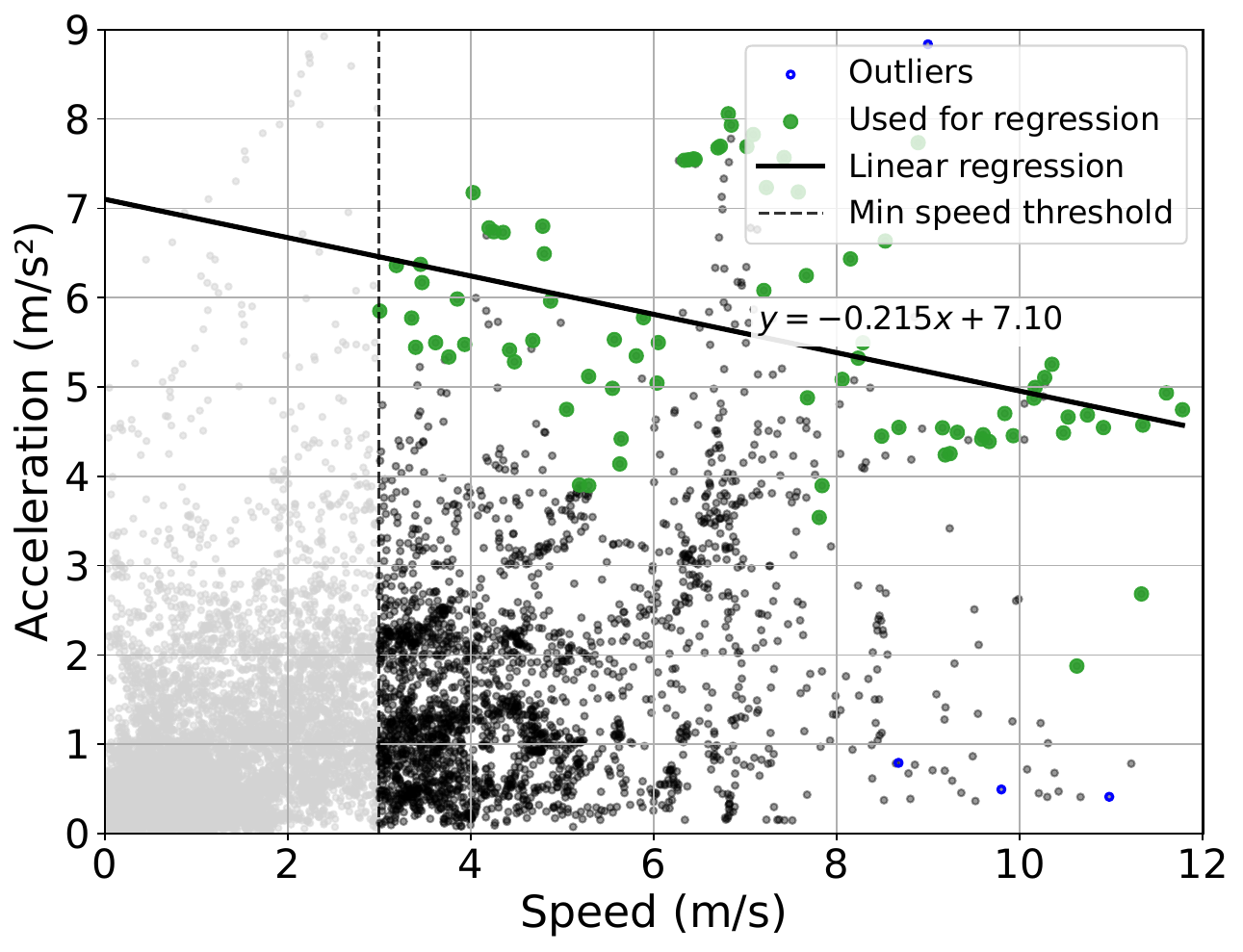}
        \caption{A--S profile derived from predicted athlete speed and acceleration signals for the first 23 minutes of the 45-minute segment.}
        \label{fig:unreliable_speed_3}
    \end{subfigure}
    \hspace{0.02\textwidth}
    \begin{subfigure}[t]{0.485\textwidth}
        \centering
        \includegraphics[width=\linewidth,trim={0 0 0 0cm},clip]{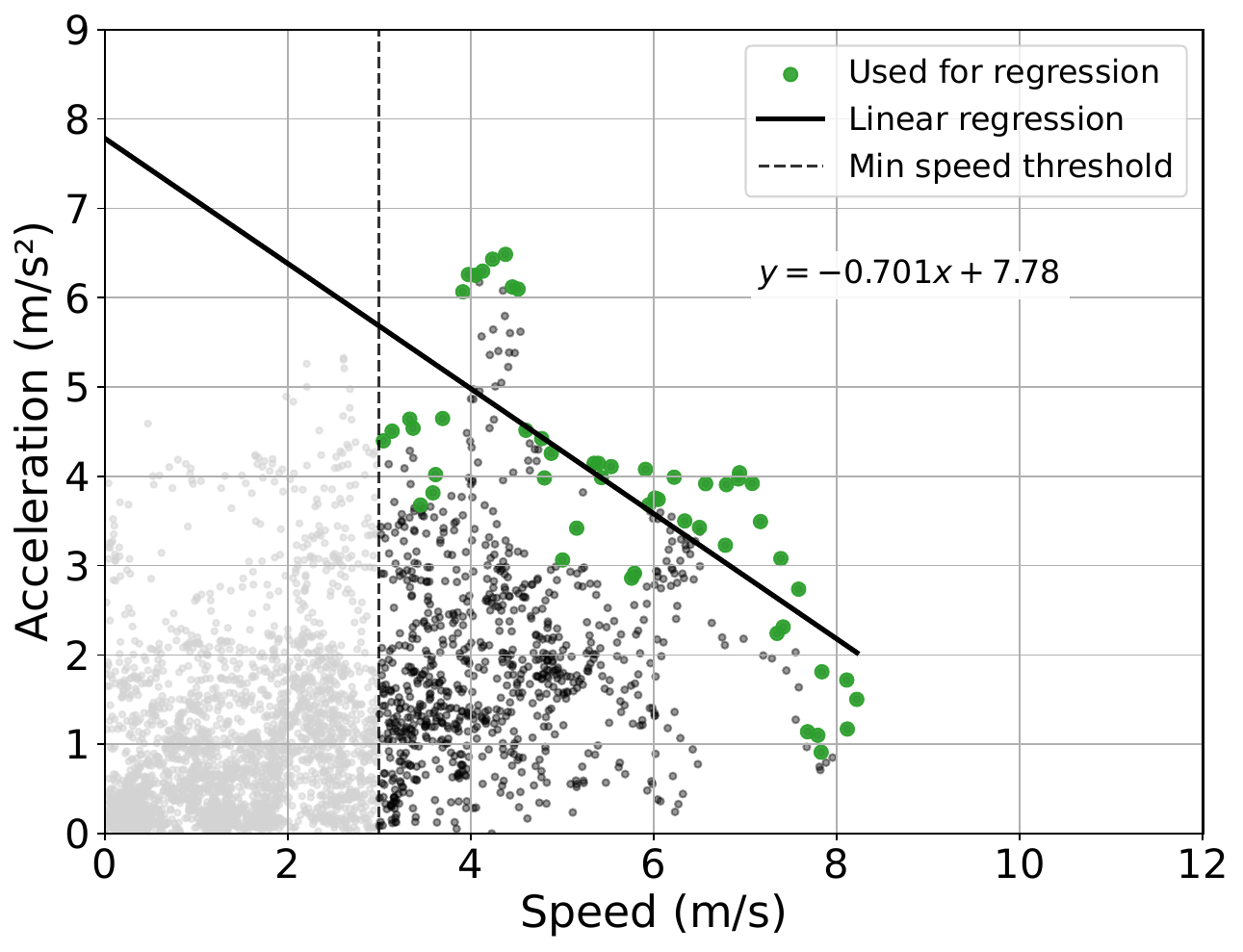}

        \caption{A--S profile derived from predicted athlete speed and acceleration signals for the second 23 minutes of the 45-minute segment.}
        \label{fig:unreliable_acc_3}
    \end{subfigure}
    \caption{\textbf{Comparison of early (left) and late (right) acceleration--speed (A--S) profiles over a 45-minute broadcast sequence.} The early profile shows a gradual decrease in acceleration as speed increases, whereas the late profile indicates a substantially faster decline.}
    \label{fig:AP_profile_SNGS-199-2}
\end{figure*}

\subsection{A--S profile results}
We compute the A--S profiles per athlete using the predicted speed and acceleration signals, as described in the previous section. \Cref{fig:AP_profile_SNGS-199} illustrates an A--S linear profile example built for an athlete of the SNGS-199 sequence. In addition, we provide a reference A--S profile derived from the ground-truth trajectory data. The theoretical maximal acceleration $A_0$ and speed $S_0$ can be extracted from the profile, \ie, $3$\,$m/s^2$ and about $9$\,$m/s$.  The absolute maximum speed and acceleration values should be interpreted with caution for the following reasons. First, although temporal correlations are strong, amplitude errors remain present, which may inflate peak values—particularly for second derivative acceleration. Secondly, the $30$-second clips are short, providing limited data points that represent the athlete's fatigue-related indicators over that particular time, rather than a generalizable individual profile.
Despite these limitations, we can use the profile to address the evolution of the A--S curve over the sequence.
Moreover, the comparison illustrates that predicted A–S profiles closely follow ground truth signals with moderate noise, demonstrating the potential for profile extraction from video.

To evaluate the method in a more realistic setting, we extract the GSR predictions of the Broadcast2Pitch method over a full 45-minute match sequence. The first and second halves are processed separately, and their corresponding A–S profiles are computed, resulting in substantially larger data samples. \Cref{fig:AP_profile_SNGS-199-2} shows the A--S profiles of one athlete for the first half (left) and the second half (right). Unfortunately, there is no ground truth reference for this example to provide a quantitative analysis. Nevertheless, comparing the early and late profiles provides qualitative insight into fatigue. In the first half, acceleration decreases gradually as speed increases. In contrast, the second-half profile exhibits a steeper slope, indicating a more rapid decline in acceleration with increasing speed. This suggests that, as the match progresses, the athlete's ability to generate force at higher speeds (e.g., during directional changes) diminishes, which is consistent with fatigue effects.

Qualitatively, both speed and acceleration signals exhibit high-amplitude fluctuations caused by noise and the first- and second-order derivatives. This issue is well documented in spatio-temporal motion estimation from broadcast video~\cite{Rustebakke2025Extracting}. Therefore, absolute amplitude values should be interpreted with caution. Instead, the relative temporal behavior of the acceleration–speed relationship is more informative for fatigue assessment. Incorporating dedicated filtering or robust estimation strategies during A–S linear model fitting could further suppress inflated values and yield more physiologically plausible profiles.

\mysection{Discussion on the reliability of ground truth and A--S profiles.} In this work, we evaluate speed and acceleration predictions derived from monocular GSR spatio-temporal outputs using publicly available datasets. %
However, ground-truth speed and acceleration are extracted from annotated athlete trajectories using a methodology similar to the one applied to predictions. This can introduce a potential bias: large $l_n$ values may oversmooth important spatio-temporal variations while improving quantitative scores, meaning evaluations could under-represent high-frequency motion dynamics.
To overcome this limitation, a GPS-based public benchmark would be highly valuable, providing precise and unbiased speed and acceleration ground truth. We encourage the scientific community to pursue this direction to enable more reliable evaluation of A--S profiles from video.

\section{Conclusion}

In this work, we investigated the prediction of fatigue-related indicators in football athletes from monocular broadcast videos. While previous studies relied on GPS data or expensive multi-camera systems, single-camera approaches offer the potential for scalable, low-cost fatigue monitoring. To this end, we developed a method to extract speed and acceleration signals from athlete trajectories predicted by state-of-the-art GSR methods, enabling the construction of acceleration–speed (A--S) profiles. Our experiments indicate that current GSR pipelines can reliably detect athletes and generate robust spatio-temporal signals for fatigue estimation. Speed predictions for the most reliable detections are particularly accurate, and acceleration predictions reported considerable scores despite higher noise from second-order derivatives. Nevertheless, errors in athlete trajectories remain common in challenging scenarios, such as occlusions or cluttered regions, which limit the reliability of derived fatigue indicators.
By relying exclusively on publicly available datasets and open-source GSR methods, our study fosters reproducible research and provides a foundation for fair, transparent comparisons in this emerging area. At the same time, the results highlight the need for strategies to identify and mitigate unreliable predictions, as well as for publicly accessible GPS-based benchmarks that provide precise, unbiased ground truth for rigorous evaluation. Overall, this work demonstrates the promise of low-cost, single-camera fatigue monitoring in football, and positions it as a compelling direction for future research in sports video understanding.

\mysection{Acknowledgments.} This research project was supported by the UEFA Medical \& Anti-Doping Research Grant Programme (UEFA MRGP) through a research grant. %

{
    \small
    \bibliographystyle{ieeenat_fullname}
    \bibliography{bib/abbreviation-short, 
    bib/all, 
    bib/PUT_NEW_REFS_HERE.bib}
}

\end{document}